\pdfoutput=1
\documentclass[11pt]{article}

\usepackage[]{acl}

\usepackage{subcaption}

\usepackage{times}
\usepackage{latexsym}
\usepackage{graphicx}
\usepackage{makecell}
\usepackage{multirow}
\usepackage{booktabs}
\usepackage{mathrsfs}

\usepackage[disable]{todonotes}
\usepackage{arydshln}

\usepackage[T1]{fontenc}
\newcommand{\metric}[1]{\textsc{#1}}
\usepackage[export]{adjustbox}
\usepackage{amsmath}
\usepackage[utf8]{inputenc}

\usepackage[T1]{fontenc}

\usepackage[utf8]{inputenc}

\usepackage{microtype}

\usepackage{amssymb}
\newcommand{\insertTableTaskPerformance}{
\begin{table*}[!htbp]
    \footnotesize
    \setlength{\tabcolsep}{3pt}
      \centering
        \begin{tabular}{l|l| rrrrr}
            \toprule
            \textbf{Settings} & \textbf{Metrics} & \textbf{Coherence} & \textbf{Consistency} & \textbf{Fluency} & \textbf{Relevance} & \textbf{Average} \\
            \midrule
            \multirow{11}{*}{$m(\mathrm{hyp}, \mathrm{ref})$}
            &\multicolumn{4}{l}{\textbf{Non-discourse metrics}}\\
            \cmidrule{2-7}
            &ROUGE-1 & 9.09 & 27.27 & 18.18 & 9.09 & 15.91\\
            &ROUGE-L & 0.00 & 36.36 & 21.21 & 18.18 & 18.94\\
            &BERTScore & 30.30 & 30.30 & 51.52 & 54.55 & 41.67 \\
            &MoverScore & 36.36 & 42.42 & 63.64 & 60.61 & 50.76 \\
            &SBERT & 3.03 & 33.33 & 30.30 & 27.27 & 23.48 \\
            &BLEURT & 45.45 & \textbf{51.52} & \textbf{72.73} & 63.64 & 58.33 \\
            & BARTScore & 60.61 & 36.36 & 45.45 & 48.48 & 47.73\\
            &PRISM & 51.52 & 39.39 & \textbf{72.73} & 69.70 & 58.33\\
            &$S^3$-pyr & 18.18 & 24.24 & 9.09 & 6.06 & 14.39 \\
            \midrule
            \multirow{6}{*}{$m(\mathrm{hyp})$}
            & \multicolumn{4}{l}{\textbf{Discourse metrics}}\\
            \cmidrule{2-7}
            &RC & 45.45 & \textbf{51.52} & \textbf{54.55} & 57.58 & \textbf{52.27} \\
            &LC & \textbf{51.52} & 45.45 & 48.48 & 57.58 & 50.76 \\
            &Entity Graph & 42.42 & 12.12 & 15.15 & 18.18 & 21.97 \\ 
            &Lexical Graph & 48.48 & 6.06 & 15.15 & 18.18 & 21.97 \\
            \midrule
            \multirow{8}{*}{$m(\mathrm{hyp}, \mathrm{ref})$}
            
            &Lexical Chain &  42.42 & 6.06 & 9.09 & 18.18 & 18.94 \\ 
            &\metric{DS-Focus} (NN) & \textbf{75.76} & \textbf{63.64} & \textbf{78.79} & \textbf{81.82} & \textbf{75.00} \\
            
            &\metric{DS-Focus} (Entity) & 69.70 & 57.58 & 72.73 & 75.76 & 68.94 \\ 
            
            &\metric{DS-Sent-u} (NN) & 48.48 & 54.55 & 63.64 & 60.61 & 56.82 \\
            &\metric{DS-Sent-u} (Entity) & 54.55 & 60.61 & 75.76 & 66.67 & 64.39 \\ 
            
            &\metric{DS-Sent-w} (NN) & 51.52 & 51.52 & 66.67 & 63.64 & 58.33 \\
            &\metric{DS-Sent-w} (Entity) &51.52 & 57.58 & 66.67 & 63.64 & 59.85\\ 
        	\bottomrule  
        \end{tabular}
   
    \caption{System-level Kendall correlations between metrics and human ratings of summary quality on SUMMEval. We bold numbers that significantly outperform others according to paired t-test~\cite{fisher1937design}. $m$ is a metric.
    \label{tab:main-task-performances} 
    }
\end{table*}
}

\newcommand{\insertTableNeR}{
\begin{table*}[!htbp]
    \footnotesize
      \centering
        \begin{tabular}{l|l| rrrrr}
            \toprule
            \textbf{Settings} & \textbf{Metrics} & \textbf{Coherence} & \textbf{Fluency} & \textbf{Informative} & \textbf{Relevance} & \textbf{Average} \\
            \midrule
            \multirow{5}{*}{$m(\mathrm{hyp}, \mathrm{ref})$}
            &BARTScore & 42.58 & 42.58 & 23.80 & 33.33 & 35.57\\
            &PRISM & 51.52 & 42.58 & 42.86 & 52.38 & 47.33\\
            &\metric{DS-Focus} (NN) & 61.90 & 61.90 & 42.86 & 52.38 & 54.76\\
            &\metric{DS-Focus*} (NN) & \textbf{80.95} & \textbf{80.95} & \textbf{100.00} & \textbf{90.47} & \textbf{88.09}\\
            
            &\metric{DS-Sent-u} (NN) & 14.29 & 14.29 & 14.29 & 23.81 & 16.67\\
        	\bottomrule  
        \end{tabular}
        \caption{System-level Kendall correlations between metrics and human ratings on NeR18. \metric{DS-Focus*} is the `F-score' version of \metric{DS-Focus}.}
        \label{tab:main-task-NeR18} 
\end{table*}
}

\newcommand{\insertTableLM}{
\begin{table}[t!]
    \footnotesize
    \centering
    \begin{tabular}{l|l | r }
    \toprule
     \textbf{Metrics} & \textbf{Encoders} & \textbf{Average} \\
    \midrule
    \multirow{3}{*}{\metric{DS-Focus} (NN)}
    &+ BERT & 71.97 \\
    &+ BERT-NLI  & 70.45 \\
    &+ Conpono  & 75.00 \\
    \midrule
    \multirow{3}{*}{\metric{DS-Sent-u} (NN)}
    & + BERT & 35.61 \\
    & + BERT-NLI & 56.82 \\
    & + Conpono & 23.48 \\
    \bottomrule
    \end{tabular}
    \caption{Results of three contextualized encoders on SUMMEval.
    Results are averaged across four aspects. 
    }
    \label{tab:LM}
\end{table}
}

\newcommand{\insertTableAblation}{
\begin{table}[t!]
    \footnotesize
    \centering
    \begin{tabular}{l | r}
    \toprule
     \textbf{Metrics} & \textbf{Average} \\
    \midrule
    \metric{DS-Sent-u} (NN) & 56.82 \\
    w/o sentence aggregation & 46.21 \\
    \bottomrule
    \end{tabular}
    \caption{Ablation study on the use of adjacency matrix to aggregate sentence embeddings on SUMMEval.
    }
    
    \label{tab:ablation}
\end{table}
}
\newcommand{\insertTableSummaryLevel}{
\begin{table}[t!]
    \footnotesize
    \centering
    \begin{tabular}{l | rr}
    \toprule
     \textbf{Metrics} & \textbf{SUMMEval} & \textbf{NeR18}\\
    \midrule
    BARTScore & 14.13 & 24.78\\
    PRISM & 14.92 & 18.89\\
    \metric{DS-Focus} (NN) & 10.81 & 10.42\\
    \metric{DS-Sent-u} (NN) & 15.71 & 3.81\\
    \bottomrule
    \end{tabular}
    \caption{Summary-level averaged Kendall correlations across all rating aspects.
    }
    
    \label{tab:summ-level}
\end{table}
}
\newcommand{\insertTableEnsemble}{
\begin{table*}[t!]
    \footnotesize
    \centering
    \begin{tabular}{l | rrrrr}
    \toprule
     \textbf{Metrics} & \textbf{Coherence} & \textbf{Consistency} & \textbf{Fluency} & \textbf{Relevance} & \textbf{Average} \\
    \midrule
    RC & 45.45 & 51.52 & 54.55 & 57.58 & 52.27\\
    BARTScore (max) & \textbf{78.79} & 48.48 & 63.64 & 72.73 & 65.91\\
    BARTScore (max) + RC & 66.67 & 54.55 & 69.70 & 78.79 & 67.42\\
    \bottomrule
    \metric{DS-Focus} (NN) & 75.76 & \textbf{63.64} & \textbf{78.79} & \textbf{81.82} & \textbf{75.00}\\
    \bottomrule
    \end{tabular}
    \caption{Ensemble of non-discourse and discourse metrics (BARTScore + RC) vs DiscoScore.}
    
    \label{tab:ensemble}
\end{table*}
}
\newcommand{\insertTableAggregation}{
\begin{table}[t!]
    \footnotesize
    \centering
    \begin{tabular}{l|l| r}
    \toprule
    \textbf{Metrics} & \textbf{Mechanisms} & \textbf{Average} \\
    \midrule
    \multirow{3}{*}{\metric{DS-Sent-u} (NN)}
    & + greedy align & 21.97 \\
    & + optimal align & 26.52 \\
    & + mean-max-min-sum & 56.82 \\
    \bottomrule
    \end{tabular}
    \caption{Averaged results of SentGraph variants based on three mechanisms on SUMMEval.
    }
    \label{tab:greedy-optimal}
\vspace{-0.05in}
\end{table}
}

\title{DiscoScore: Evaluating Text Generation \\ with 
    BERT and Discourse Coherence}
  
\author{Wei Zhao$^{1 2}$ \text{ } Michael Strube$^{1}$ \text{ }
Steffen Eger$^{3}$ \\
    $^{1}$Heidelberg Institute for Theoretical Studies
    \text{ } $^2$Technische Universit\"at Darmstadt \\
    \url{www.h-its.org/research/nlp/} \\
    $^3$NLLG, Faculty of Technology, Bielefeld University \\
    {\tt \{wei.zhao, michael.strube\}@h-its.org}\\
    {\tt steffen.eger@uni-bielefeld.de}\\
    \url{nl2g.github.io}
  }

\begin{document}
\maketitle
\begin{abstract}

Recently, there has been a growing interest in designing text generation systems from a discourse coherence perspective, e.g., modeling the interdependence between sentences. Still, recent BERT-based evaluation metrics are weak in recognizing coherence, and thus are not reliable in a way to spot the discourse-level improvements of those text generation systems.
In this work, we introduce DiscoScore, a parametrized discourse metric, which uses BERT to model discourse coherence from different perspectives, driven by Centering theory. Our experiments encompass 16 non-discourse and discourse metrics, including DiscoScore and popular coherence models, evaluated on summarization and document-level machine translation (MT). We find that (i) the majority of BERT-based metrics correlate much worse with human rated coherence than early discourse metrics, invented a decade ago; (ii) the recent state-of-the-art BARTScore is weak when operated at system level---which is particularly problematic as systems are typically compared in this manner. DiscoScore, in contrast, achieves strong system-level correlation with human ratings, not only in coherence but also in factual consistency and other aspects, and surpasses  BARTScore by over 10 correlation points on average.  Further, aiming to understand DiscoScore, we provide justifications to the importance of discourse coherence for evaluation metrics, and explain the superiority of one variant over another. Our code is available at \url{https://github.com/AIPHES/DiscoScore}.
\end{abstract}

\section{Introduction}
In discourse, coherence refers to the continuity of semantics in text. Often, discourse relations and lexical cohesion devices, such as repetition and coreference, are employed to connect text spans, aiming to ensure text coherence.
Popular theories in the linguistics community on discourse 
were provided by \citet{grosz-etal-1995-centering} and \citet{mann1988rhetorical}. They formulate coherence through the lens of readers' focus of attention, and rhetorical discourse structures over sentences.
Later on, coherence models as computational approaches of these theories emerged to judge text coherence in discourse tasks such as sentence ordering and essay scoring~\cite{barzilay2008modeling, lin-etal-2011-automatically, guinaudeau-strube-2013-graph}.

While humans also often use text planning at discourse level 
prior to writing and speaking, 
up until recently, the majority of 
natural language generation (NLG) systems, 
be it text summarization or document-level MT, has performed sequential word prediction without considering text coherence. For instance, MT systems mostly do not model the interdependence between sentences and translate a document at sentence level, and 
thus 
produce many incoherent elements such as coreference mistakes in system outputs~\cite{maruf2021survey}.
Only more recently 
has there been a surge of interest 
towards discourse based summarization and MT systems, aiming to model
inter-sentence context, with a focus on pronominal anaphora~\cite{voita-etal-2018-context, liu-etal-2021-coreference} and discouse relations~\cite{miculicich-etal-2018-document, xu-etal-2020-discourse}. 

However, there appears a mismatch between discourse based NLG systems and non-discourse NLG evaluation metrics such as MoverScore~\cite{zhao-etal-2019-moverscore} and BERTScore~\cite{DBLP:conf/iclr/ZhangKWWA20} 
which have recently become popular for MT and summarization evaluation. 
As these metrics base their judgment on semantic similarity (and lexical overlap \citep{kaster-etal-2021-global}) between hypotheses and references---which by design does not target text coherence---it is not surprising that they do not correlate well with human rated coherence~\cite{fabbri2021summeval, yuan2021bartscore, sai-etal-2021-perturbation}. Recently, BARTScore~\cite{yuan2021bartscore} receives increasingly attention, which uses sequence-to-sequence language models to measure the likelihood that hypothesis and reference are paraphrases, and that cannot contrast text pairs at discourse level.

\begin{figure}
    \begin{subfigure}{0.5\textwidth}
    \centering
    \includegraphics[width=0.9\textwidth]{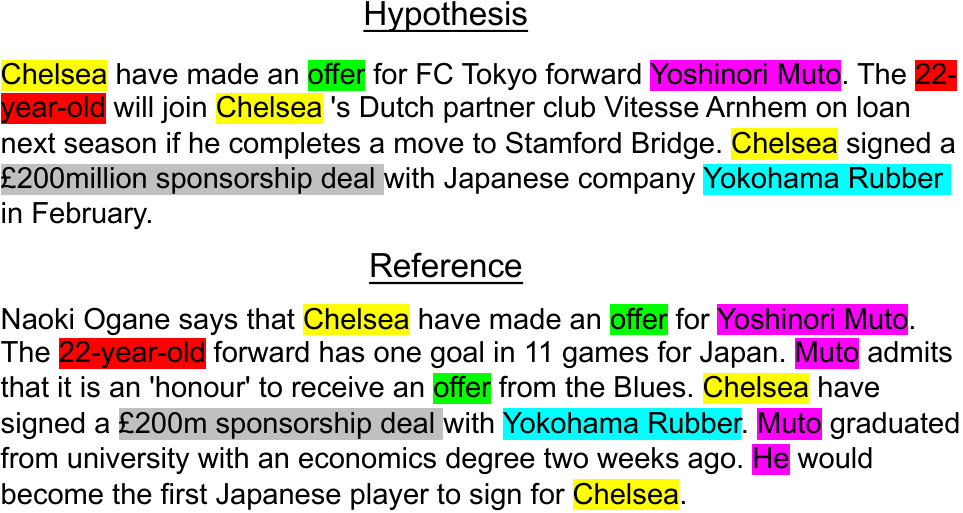}
    \end{subfigure}\par\vskip\floatsep
    \begin{subfigure}{.59\linewidth}
      \centering
            \resizebox{\columnwidth}{!}{%
            \begin{tabular}{ l| c c c c c c}
            \footnotesize
                   & $t_1$ & $t_2$ & $t_3$ & $t_4$ & $t_5$ & ... \\
                  \hline
                  Chelsea & 1 & 0 & 0 & 0 & 0 & 1 \\
                  offer & 0 & 0 & 0 & 0 & 1 & 0\\
                  $\vdots$ & $\vdots$ & $\vdots$ & $\vdots$ & $\vdots$ & $\vdots$ & $\vdots$ \\
            \end{tabular}
            }
            \caption{FocusDiff}
    \end{subfigure}%
    \hfill  
    \begin{subfigure}{.41\linewidth}
      \centering
        \resizebox{0.8\columnwidth}{!}{%
        \begin{tabular}{ l| c c c}
               & $s_1$ & $s_2$ & $s_3$ \\
              \hline
              $s_1$ & 0 & 1 & 0.5 \\
              $s_2$ & 0 & 0 & 1 \\
              $s_3$ & 0 & 0 & 0 \\
        \end{tabular}
        }
        \caption{SentGraph}
    \end{subfigure} 
    \caption{
    Sample hypothesis and reference from SUMMEval. 
    Each focus\protect\footnotemark is marked in a different color, corresponding to multiple tokens as instances of a focus. Foci shared in Hypothesis and Reference are marked in the same color. (a)+(b) are adjacency matrices used to model focus-based coherence for Hypothesis; for simplicity, adjacency matrices for Reference are omitted. FocusDiff and SentGraph are the variants of DiscoScore.
    For FocusDiff, we use (a) to depict the relations between foci and tokens, reflecting focus frequency. For SentGraph, we use (b) to depict the interdependence between sentences according to the number of foci shared between sentences and the distance between sentences.}
    \label{fig:example}  
\end{figure}

In this work, we fill the gap of 
missing 
discourse metrics in MT and summarization evaluation, particularly in reference-based evaluation scenarios. We introduce DiscoScore, a parametrized discourse metric, 
which uses BERT to model discourse coherence through the lens of readers' focus, driven by Centering theory~\cite{grosz-etal-1995-centering}. The DiscoScore variants can be distinguished in how we use \emph{focus}---see Figure~\ref{fig:example}: (i) we model focus frequency and semantics, and compare their difference between hypothesis and reference and (ii) we use focus transitions to model the interdependence between sentences. Building upon this, we present a simple graph-based approach to compare hypothesis with reference.

\footnotetext{The formal definition of focusing in discourse is given on two levels \cite{grosz1977representation}: (i) readers are said to be \textit{globally} focusing on a set of entities relevant to the overall discourse, and (ii) readers focus on a particular entity that an utterance \textit{locally} concerns most. Section~\ref{sec:approach} elaborates on focus as a key ingredient of DiscoScore.}

We compare DiscoScore with a range of baselines, including discourse and non-discourse metrics, and coherence models on summarization and document-level MT datasets.
Our contributions and findings are summarized as follows: 
\begin{itemize}

    \item Recent BERT-based metrics and the state-of-the-art BARTScore~\cite{yuan2021bartscore} are all weak in system-level correlation with human ratings, not only in coherence but also in other aspects such as factual consistency. 
    Most of them are even worse than very early discourse metrics, RC and LC~\cite{wong-kit-2012-extending}---which require neither source texts nor references and use discourse features
    to predict hypothesis coherence. 
    
    \item DiscoScore strongly correlates with human rated coherence and many other aspects, over 10 points (on average across aspects) better than BARTScore and two strong baselines RC and LC in the single and multi-references settings. This indicates that either leveraging contextualized encoders or finding discourse features is not sufficient, suggesting to combine both as DiscoScore does.
    
    \item We demonstrate the importance of including discourse signals in the assessment of system outputs, as the discourse features derived from DiscoScore can strongly separate hypothesis from reference. Further, we show that the more discriminative these features are, the better the metrics perform, 
    which allows for interpreting the performance gaps between the variants of DisoScore.
     
    \item We investigate two focus choices popular in the discourse community, i.e., noun~\cite{elsner2011extending} and semantic entity~\cite{mesgar-strube-2016-lexical}. Our results show that entity as focus is not always helpful, but when it helps, the gain is big.
    
\end{itemize}

\section{Related work}
\paragraph{Evaluation Metrics.}
Traditional metrics such as BLEU~\cite{Papineni:2002} and ROUGE~\cite{Lin:2004} measure lexical n-gram overlap between a hypothesis and a human reference. 
As they fail to measure semantic similarity 
in the absence of lexical overlap, 
several metrics have been proposed to overcome this issue,
which carry out soft lexical matching
with static word embeddings~\cite{ng-abrecht-2015-better} and synonym matching~\cite{lavie-agarwal-2007-meteor}. 
However, none of those metrics can properly judge text coherence~\cite{kryscinski-etal-2019-neural, zhu-bhat-2020-gruen}.

Recently, a class of novel metrics based on BERT~\cite{devlin-etal-2019-bert} has received a surge of attention, as they correlate strongly with human judgment of text quality in both reference-based and reference-free scenarios~\cite{zhao-etal-2019-moverscore,DBLP:conf/iclr/ZhangKWWA20, sellam-etal-2020-bleurt, rei-etal-2020-comet, gao-etal-2020-supert, thompson-post-2020-automatic, zhao-etal-2020-limitations, pu-etal-2021-learning, chen-etal-2021-training}. 
While strong at sentence-level, these metrics 
are weak in recognizing coherence in inter-sentence contexts (just like BLEU and ROUGE), as BERT and the majority of BERT variants\footnote{Recently, several discourse BERT variants such as Conpono~\cite{iter-etal-2020-pretraining} have been proposed, but they are not always helpful for evaluation metrics---see Table~\ref{tab:LM} (appendix).} that these metrics build on only capture discourse phenomena to a certain extent~\cite{koto-etal-2021-discourse, laban-etal-2021-transformer, beyer-etal-2021-incoherence}. Thus, they are not suitable for evaluating long texts as in document-level MT evaluation. 
Works that 
either (i) average sentence-level evaluation scores as document score or (ii) assign a score to the concatenation of sentences within a document ~\cite{xiong2019modeling, liu2020multilingual, saunders-etal-2020-using}
do not factor interdependence between sentences into a document score, e.g., do not explicitly punish incoherent elements, 
thus are also inadequate.

Several attempts have been made towards discourse metrics in MT evaluation. \citet{wong-kit-2012-extending, gong2015document, Cartoni:2018} use the frequency of lexical cohesion devices (e.g., word repetition) over sentences to predict coherence of hypothesis translations, while \citet{guzman-etal-2014-using} and \citet{joty2017discourse} suggest to compare the difference of rhetorical structures between hypothesis and reference translations. Recently, \citet{blond:2021} measure the inconsistency between hypothesis and reference translations in several aspects such as verb tense and named entities. However, 
these metrics 
do not leverage strong contextualized encoders, as has been shown to be
a key ingredient for recent success of BERT-based metrics. Most recently, BARTScore~\cite{yuan2021bartscore} uses sequence-to-sequence pretrained language models such as BART~\cite{lewis-etal-2020-bart} to measure how likely hypothesis and reference are paraphrased according to the probability of one given the other. While BARTScore
constitutes the recent state-of-the-art in sentence-level correlation with human ratings in several aspects (incl. discourse), we find that (i) it performs still poorly 
at system level---which is particularly problematic as systems are typically compared in this manner. 
(ii) As based on 
a `blackbox' language model, it cannot offer insights towards how it models coherence and what discourse phenomena it does (not) capture.

\paragraph{Coherence Models.}In discourse, there have been many computational models~\cite{barzilay2008modeling, guinaudeau-strube-2013-graph, pitler-nenkova-2008-revisiting, lin-etal-2011-automatically} for 
text coherence assessment, the majority of which differ in \textit{regularities} that they use to distinguish coherent from incoherent text, driven by different linguistic theories, \emph{v.i.z.}, a pattern of (i) focus transitions in adjacent sentences~\cite{grosz-etal-1995-centering}
and (ii) text organization regarding discourse relations over sentences~\cite{mann1988rhetorical}. For instance, \citet{barzilay2008modeling} and \citet{guinaudeau-strube-2013-graph} use the distribution of entity transitions over sentences to predict text coherence, while \citet{pitler-nenkova-2008-revisiting} and \citet{lin-etal-2011-automatically} suggest to produce discourse relations over sentences with a discourse parser, showing that the relations are indicative of text coherence. 
In the last few years, neural coherence models have been explored.
Popular examples are \citet{tien-nguyen-joty-2017-neural}, \citet{mesgar-strube-2018-neural} and \citet{moon-etal-2019-unified}. As they and the recent state-of-the-art~\cite{mesgar-etal-2021-neural-graph} all have been trained on text readability datasets, with readability labels as supervision, they may suffer issues of domain shift when applied to MT and summarization evaluation.
More importantly, they judge hypothesis coherence in the absence of reference, thus are not sufficient for reference-based evaluation. Our experiments involve two popular, unsupervised coherence models, entity graph~\cite{guinaudeau-strube-2013-graph} and lexical graph~\cite{mesgar-strube-2016-lexical} treated as discourse metrics with the advantages on robustness~\cite{lai-tetreault-2018-discourse}.
   
\paragraph{Discourse Test Sets.} Apart from evaluation metrics, there have been several discourse-focused test sets proposed to compare NLG systems, most of which have been studied in MT evaluation. For instance, the  DiscoMT15 shared task~\cite{hardmeier-etal-2015-pronoun} compares MT systems, not based on translation adequacy but on the accuracy of pronoun translation for English-to-French, i.e., counting the number of correctly translated pronouns, given the annotated ones in reference.
\citet{bawden-etal-2018-evaluating} extend this by labeling both anaphoric pronouns and lexical cohesion devices on test sets, while \citet{voita-etal-2018-context} construct English-to-Russian test sets focusing on deixis, ellipsis and lexical cohesion. \citet{guillou-etal-2018-pronoun, lopes:2020} construct  English-to-German and English-to-French test sets targeting pronouns. While reliable, these test sets involve costly manual annotation, thus are limited to few language pairs.

In this work, we introduce DiscoScore to judge system outputs, which uses BERT to model readers' focus within hypothesis and reference, and thus clearly outlines the discourse phenomena being captured,
serving as low-cost alternatives to discourse test sets for comparing discourse based NLG systems. More prominently, we derive discourse features from DiscoScore, which we use to understand the importance of discourse for evaluation metrics, and explain why one metric is superior to another. This parallels 
recent effort towards explainability for non-discourse evaluation metrics~\cite{kaster-etal-2021-global, fomicheva-etal-2021-eval4nlp}. 
Finally, we show that simple features can be indicative of the superiority of a metric, which fosters research towards finding insightful features with domain expertise and building upon these insights to design high-quality metrics.

\section{Our Approach}
\label{sec:approach}
In the following, we elaborate on the two variants of DiscoScore, FocusDiff and SentGraph, 
which we refer to 
as \metric{DS-Focus} and \metric{DS-Sent}.

\paragraph{Focus Difference.} In discourse, there have been many 
corpus-based studies towards 
modeling focus transitions over sentences,
showing that  focus transition patterns are indicative of text coherence~\cite{barzilay2008modeling,guinaudeau-strube-2013-graph}. When reading a document, readers may 
have multiple \emph{focus of attention}, 
 
each associated to a group of expressions: (i) referring expressions such as pronouns and (ii) semantically related elements such as [\textit{Berlin, capital}].

Here, we 
assume two focus based conditions that a coherent hypothesis should meet in reference-based evaluation scenarios:
\begin{itemize}
    \item A large number of focus overlaps between a hypothesis and a reference. 
    \item Each focus overlap is nearly identical in terms of semantics and frequency, where frequency shows how often a focus is mentioned in a hypothesis or in a reference.
\end{itemize}

In the following, we present focus modeling towards semantics and frequency, according to which we compare hypothesis with reference.

For a hypothesis, we introduce a bipartite graph $\mathcal{G}^{\rm{hyp}}=(\mathcal{V}, \mathcal{S}, \mathbf{A}^{\rm{hyp}})$, where $\mathcal{V}$ and $\mathcal{S}$ are two sets of vertices corresponding to a set of foci and all tokens (per occurrence a word is a separate token) within a hypothesis. Let $\mathbf{A}=\{0, 1\}^{n\times m}$ be an adjacency matrix where $n$ and $m$ are the number of foci and tokens respectively, and $A_{ij}$ equals $1$ if and only if the $i$-th focus associates to the $j$-th token.  
Let $\mathbf{F}^{\rm{hyp}}\in \mathbb{R}^{n\times d}$ be a matrix of focus embeddings and $\mathbf{Z}^{\rm{hyp}}\in \mathbb{R}^{m\times d}$ be a matrix of contextualized token embeddings with $d$ as the embedding size. Similarly, we use notation $\mathcal{G}^{\rm{ref}}$, $\mathbf{F}^{\rm{ref}}$ and $\mathbf{Z}^{\rm{ref}}$ for a human reference.

We use contextualized encoders such as BERT
to produce token embeddings $\mathbf{Z}^{\rm{hyp}}$ and $\mathbf{Z}^{\rm{ref}}$. We use a simple approach to model both semantics and frequency of a focus. That is, we assign  per focus $v$ an embedding by summing token embeddings that a focus is associated to:
\begin{equation}
\label{eq:vector-form}
    \mathbf{F}^{\rm{hyp}}_v = \sum_{u \in \mathcal{N}(v)} \mathbf{Z}^{\rm{hyp}}_u ,\;
    \mathbf{F}^{\rm{ref}}_v = \sum_{u \in \mathcal{N}(v)} \mathbf{Z}^{\rm{ref}}_u 
\end{equation}  
where $\mathcal{N}(v)$ is a set of tokens (e.g., a group of semantically related expressions) associated with a focus $v$. In matrix notation, we rewrite Eq.~(\ref{eq:vector-form}) to $\mathbf{F}^{\rm{hyp}} = \mathbf{A}^{\rm{hyp}}\mathbf{Z}^{\rm{hyp}}$, similarly for $\mathbf{F}^{\rm{ref}}$.

Next, we measure the distance between a common set of foci $\Omega$ in a hypothesis and reference pair based on their embeddings: 
	 \begin{equation}
	 \label{eq:penalty}
	        \metric{DS-Focus}(\rm{hyp}, \rm{ref}) = \frac{1}{N}\sum_{u \in \Omega}\| \mathbf{F}^{\rm{hyp}}_{u} - \mathbf{F}^{\rm{ref}}_{u} \|
	 \end{equation}
	 where 
	 \metric{DS-Focus} is scaled down by the factor of $N$, the number of foci in hypothesis.
    
\paragraph{Sentence Graph.}

    Few contextualized encoders can produce high-quality sentence embeddings in the document context, as they do not model interdependence between sentences.  
    According to Centering theory~\cite{grosz-etal-1995-centering}, two sentences are marked continuous in meaning when they share at least one focus, on the one hand; one marks a meaning shift for two sentences when no focus appears in common, on the other hand. From this, one can aggregate sentence embeddings for which corresponding sentences are considered continuous. In the following, we present a graph-based approach to do so.

    For a hypothesis\footnote{For simplicity, we omit the notation $\mathbf{S}^{\rm{ref}}$ and $\mathcal{G}^{\rm{ref}}$ for a reference.}, let $\mathbf{S}^{\rm{hyp}}\in \mathbb{R}^{n\times d}$ be a matrix of sentence embeddings with $n$ and $d$ as the number of sentences and the embedding size.
    We introduce a graph $\mathcal{G}^{\rm{hyp}}=(\mathcal{V}, \mathbf{A}^{\rm{hyp}})$ where $\mathcal{V}$ is a set of sentences and $\mathbf{A}^{\rm{hyp}}$ is an adjacency matrix weighted according to the number of foci shared between sentences and the distance between sentences as listed below to depict two variants of $\mathbf{A}^{\rm{hyp}}$: 
    \begin{itemize}
        \item unweighted: 
        $\mathbf{A}^{\rm{hyp}}_{ij}=1/(j-i)$ if the $i$-th and the $j$-th sentences have at least one focus in common (otherwise 0), where $j-i$ denotes the distance between two sentences and $\mathbf{A}^{\rm{hyp}}_{ij}=0$ when $j\leq i$. 
        \item weighted: $\mathbf{A}^{\rm{hyp}}_{ij}=a/(j-i)$, where $a$ is the number of foci shared in the $i$-th and the $j$-th sentences, with the same constraints on $j$ and $i$ as above. 
    \end{itemize}
    
    Analyses by \citet{guinaudeau-strube-2013-graph} indicate that global statistics (e.g., average) over such adjacency matrices can distinguish incoherent from coherent text to some degree. Here we depict adjacency matrices as a form of sentence connectivity derived from focus transitions over sentences. We use them to aggregate sentence embeddings from hypothesis and from reference: 
    
    \begin{equation}
    \label{eq:sent-graph}
        \mathbf{\hat{S}}^{\rm{hyp}} = (\mathbf{A^{\rm{hyp}}}+\mathbf{I})\mathbf{S}^{\rm{hyp}},\;
        \mathbf{\hat{S}}^{\rm{ref}} = (\mathbf{A^{\rm{ref}}}+\mathbf{I})\mathbf{S}^{\rm{ref}}
        \nonumber
    \end{equation}  
    where $\mathbf{I}$ is an identity matrix that adds a self-loop to a graph so as to include self-embeddings when updating them. 
    
    Next, we derive per graph an embedding with simple statistics from $\mathbf{\hat{S}}^{\rm{hyp}}$ and $\mathbf{\hat{S}}^{\rm{ref}}$, i.e., the concatenation of mean-max-min-sum embeddings.
    Finally, we compute the cosine similarity between two graph-level embeddings:
    \begin{equation}
	        \metric{DS-Sent}(\rm{hyp}, \rm{ref}) = \mathrm{cosine}(\mathcal{G}^{\mathrm{hyp}}, \mathcal{G}^{\mathrm{ref}})
	 \end{equation}

	\paragraph{Choice of Focus.} In discourse, often four popular choices are used to describe a focus: (i) a noun that heads a NP~\cite{barzilay2008modeling}, (ii) a noun~\cite{elsner2011extending}, (iii) a coreferent entity associated with a set of referring expressions~\cite{guinaudeau-strube-2013-graph} and (iv) a semantic entity associated with a set of lexical related words~\cite{mesgar-strube-2016-lexical}.
	
	In this work, we investigate two focus choices: noun (NN) and semantic entity (Entity).
	Linguistically speaking, the latter is a lexical cohesion device in the form of repetition.
    From this, NN as focus yields few useful coherence signals but a lot of noise, while Entity as focus uses `signal compression' by means of aggregation to produce better signals. 
	To produce entities, we first extract all nouns in hypothesis (or reference), and aggregate them into different semantic entities if their cosine similarities based on Dep2Vec word embeddings~\cite{levy2014dependency} is greater than a threshold---assuming that nouns with high similarity refer to the same semantic entity.

\section{Experiments}
\subsection{Evaluation Metrics}
In the following, we list all of the evaluation metrics, and elaborate on them in Appendix~\ref{sec:details}.
\paragraph{Non-discourse Metrics.} We consider BLEU~\cite{Papineni:2002}, ROUGE~\cite{Lin:2004}, BERTScore~\cite{DBLP:conf/iclr/ZhangKWWA20},  MoverScore~\cite{zhao-etal-2019-moverscore}, SBERT~\cite{DBLP:conf/emnlp/ReimersG19}, $S^{3}$-pyr~\cite{peyrard-etal-2017-learning}, BLEURT~\cite{sellam-etal-2020-bleurt}, BARTScore~\cite{yuan2021bartscore},  PRISM~\cite{thompson-post-2020-automatic}.

\paragraph{Discourse Metrics.} We consider RC and LC~\cite{wong-kit-2012-extending} and Lexical Chain~\cite{gong2015document}. We consider two coherence models, EntityGraph~\cite{guinaudeau-strube-2013-graph} and LexicalGraph~\cite{mesgar-strube-2016-lexical}, and treat them as discourse metrics.

\paragraph{DiscoScore.} \metric{DS-Focus} can be parameterized with two focus choices: noun (NN) or semantic entity (Entity). \metric{DS-Sent} can be parameterized not only with focus, but also with the choices of \emph{unweighted} (-U) and \emph{weighted} (-W).
For \metric{DS-Focus}, we use Conpono~\cite{iter-etal-2020-pretraining} that finetuned BERT with a novel discourse-level objective regarding sentence ordering. For \metric{DS-Sent}, we use BERT-NLI. This is because we find this configuration performs best after initial trials---see Table~\ref{tab:LM} (appendix). Figure~\ref{fig:mapping} (appendix) shows all variants of DiscoScore. Concerning the threshold of Dep2Vec to produce entities, after experimenting with several alternatives we set it to 0.8 for \metric{DS-Focus} (Entity) in all setups, and to 0.8 in summarization and to 0.5 in MT for \metric{DS-Sent} (Entity).

\subsection{Datasets}

We consider two datasets in summarization: SummEval~\cite{fabbri2021summeval} and NeR18~\cite{grusky-etal-2018-newsroom}, and one dataset in document-level MT: WMT20~\cite{mathur-etal-2020-results}. Note that these datasets consist of hypotheses paired with human-written references, where hypotheses are machine-generated texts of varying qualities given by neural and non-neural, extractive and abstractive language models.
We outline these datasets in Appendix \ref{sec:datasets}, and provide data statistics in Table~\ref{tab:statistics} (appendix).

\section{Results}
We first examine the importance of discourse for evaluation metrics---which underpins the usefulness of discourse metrics,
and then benchmark DiscoScore on summarization and MT datasets.

\paragraph{Importance of Discourse.}
\metric{DS-Focus} and \metric{DS-Sent} concern the modeling of discourse coherence on two different levels: (i) the occurrences of foci, and (ii) the interdependence between sentences driven by focus transitions, both reflecting the discourse characteristics of a text. In the following, we describe these discourse features, and examine their importance
for assessing system outputs by contrasting the discourse patterns of hypothesis and reference.
 
\begin{itemize}
    \item \textbf{Focus Frequency}, denoted by \metric{FREQ}($x$), equals the ratio between the total frequencies of foci and the number of foci in a text $x$, where $x$ is hypothesis or reference. We exclude foci occurring only once.

    \item \textbf{Sentence Connectivity}, denoted by \metric{Conn}($x$), equals the average of all elements in adjacency matrix representing the interdependence between sentences in a text $x$ (hypothesis/reference).
    
    \item As in DiscoScore, we consider two focus choices (NN and Entity) and the choices of \emph{unweighted} (-U) and \emph{weighted} (-W) for these discourse features. Figure~\ref{fig:mapping} (appendix) shows the links between DiscoScore and the features.
\end{itemize}

\begin{figure}
\centerline{\includegraphics[width=0.7\linewidth]{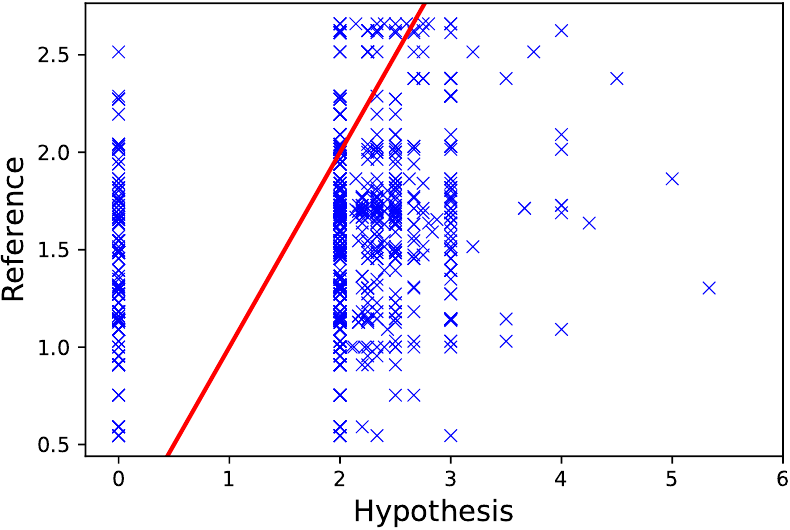}}  
 \caption{Scatter plot to display \metric{FREQ}(hyp) (based on NN) on x-axis and \metric{FREQ}(ref) on y-axis on SUMMEval. Each point contains two frequencies from a pair of hypothesis and reference.
 The points below the auxiliary line are the ones for which \metric{FREQ}(hyp) > \metric{FREQ}(ref).
 }
 \label{fig:freq-nn}  
\end{figure}

Figure~\ref{fig:freq-nn} shows that the scales on \metric{FREQ}(ref) and \metric{FREQ}(hyp) in summarization differ by a large amount, i.e., from 0.5 to 2.5 on y-axis and up to 6 on x-axis. This means that hypothesis and reference can be strongly distinguished by \metric{Freq}(x), which underpins the usefulness of including such discourse signals in the assessment of system outputs when references are available. Further, the larger scale on \metric{FREQ}(hyp) indicates that foci in hypothesis are more repetitive than in reference, as a result of needless repetition in poor summaries---in line with previous studies on incoherent machine translations~\cite{guillou-2013-analysing, voita-etal-2019-good}.
The results for other discourse features are similar, we provide them in Figure~\ref{fig:dist-all} (appendix). 

Overall, these results show discourse features can separate hypothesis from reference.

\subsection{Text Summarization}

\insertTableTaskPerformance

\paragraph{Correlation Results.}
Table~\ref{tab:main-task-performances} compares metrics on SUMMEval on system level. Most of non-discourse metrics have a lowest correlation with human rated coherence among four quality aspects. Even worse, ROUGE-L and SBERT do not correlate with coherence whatsoever.
BARTScore, the recent state-of-the-art metric, is very weak when operated on system level, 
notwithstanding that it has been fine-tuned on ``document-to-summary'' parallel data from CNN/DailyMail---which 
SUMMEval 
is 
constructed 
from.
We note that SUMMEval uses multiple references. BARTScore by default compares a hypothesis with one reference at a time, then takes the average of multiple evaluation scores as a final score. Table~\ref{tab:main-extra-SUMMEval} (appendix) shows that we can improve system-level BARTScore to some degree by replacing `average' with `max' (i.e., taking the maximum score), but \metric{DS-Focus} is still much better overall, i.e., surpassing BARTScore by ca. 10 points on average. 

Table~\ref{tab:main-task-NeR18} (appendix) reports correlation results on NeR18 that uses single reference.
We find that half of hypotheses do not contain `good foci', and as such the foci-based discourse features outlined previously are less discriminative on NeR18 than on SUMMEval---see Table~\ref{tab:statistics} (appendix). 
However, \metric{DS-Focus} is still strong, ca. 20 points better than BARTScore in all aspects, despite that \metric{DS-Focus} uses a much smaller contextualized encoder\footnote{\metric{DS-Focus} uses Conpono on the same size of BERTBase. BARTScore uses BARTLarge finetuned on CNN/DailyMail.}.
We note that the `F-score' version of \metric{DS-Focus} seems extremely strong on NeR18, but it is not robust across datasets, e.g., much worse than the original, precision-based \metric{DS-Focus} on SUMMEval.

\begin{figure}
\centerline{\includegraphics[width=0.8\linewidth]{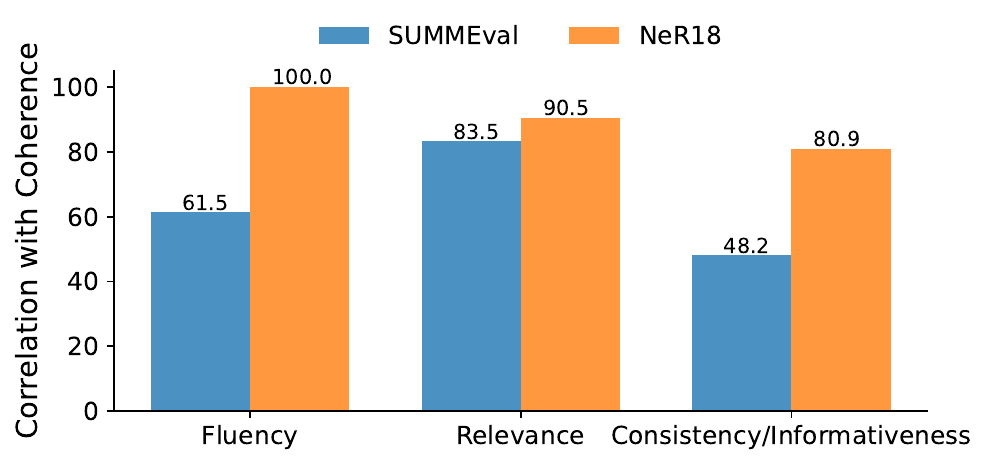}}  
 \caption{Pearson Correlation between coherence and other aspects on system level. SUMMEval and NeR18 use Consistency and Informativeness respectively.}
 \label{fig:rating_corr}  
\end{figure}

On a side note, 
coherence (mostly) strongly correlates with the other rating aspects on both SUMMEval and NeR18---see Figure~\ref{fig:rating_corr}. Thus, it is not surprising that both \metric{DS-Focus} and \metric{DS-Sent} correlate well with these aspects, despite that we have not targeted them. While strong on system level, DiscoScore could not show advantages on summary level---see Table~\ref{tab:summ-level} (appendix), but we argue that system-level correlation deserves the highest priority as systems are compared in this manner.

Overall, these results show that BERT-based non-discourse metrics correlate weakly with human ratings on system level. BARTScore also does so, though we improve it to some degree in multi-references settings. DiscoScore, particularly \metric{DS-Focus}, performs consistently best in both single- and multi-references settings, and it is equally strong in all aspects.

As for discourse metrics, RC and LC that use discourse features are strong baselines as they outperform most of non-discourse metrics and coherence models (i.e., Entity and Lexical Graph) without the access to source texts and references. However, they are worse than both \metric{DS-Focus} and \metric{DS-Sent}. This confirms the inadequacy of RC and LC in that they do not leverage strong contextualized encoders and judge hypothesis in the absence of references. Moreover, we compare DiscoScore to a combination 
of two strong, complementary baselines, BARTScore and RC---a simple solution to address text coherence of non-discourse metrics. To combine them, we simply average their scores. We see the gains are additive in all aspects but coherence. \metric{DS-Focus} wins all the time by a large margin---see Table~\ref{tab:ensemble} (appendix).

Taken together, these results show that any of the three---(i) leveraging contextualized encoders as in BERT-based metrics and BARTScore; (ii) leveraging discourse features as in RC and (iii) the ensemble of (i) and (ii) by averaging---is not sufficient, suggesting to combine (i) and (ii) as DiscoScore does.

\paragraph{Understanding DiscoScore.} As for all variants of DiscoScore, we provide understanding on why one variant is superior to another with the discourse features outlined in Figure~\ref{fig:mapping} (appendix). To this end, we begin with defining the \emph{discriminativeness} of these features as the magnitude of separating hypothesis from reference:
\begin{equation}\label{eq:pref}
    \mathcal{D}_{\mathcal{R}}(\mathrm{hyp},\mathrm{ref}):=\frac{|\{(\mathrm{hyp},\mathrm{ref})|\mathcal{R}(\mathrm{ref}) < \mathcal{R}(\mathrm{hyp})\}|}{N}
\end{equation}
where $N$ is a normalization term, $\mathcal{R}$ is any one of the discourse features in Figure \ref{fig:mapping} (appendix).

\begin{figure}
\centerline{\includegraphics[width=0.8\linewidth]{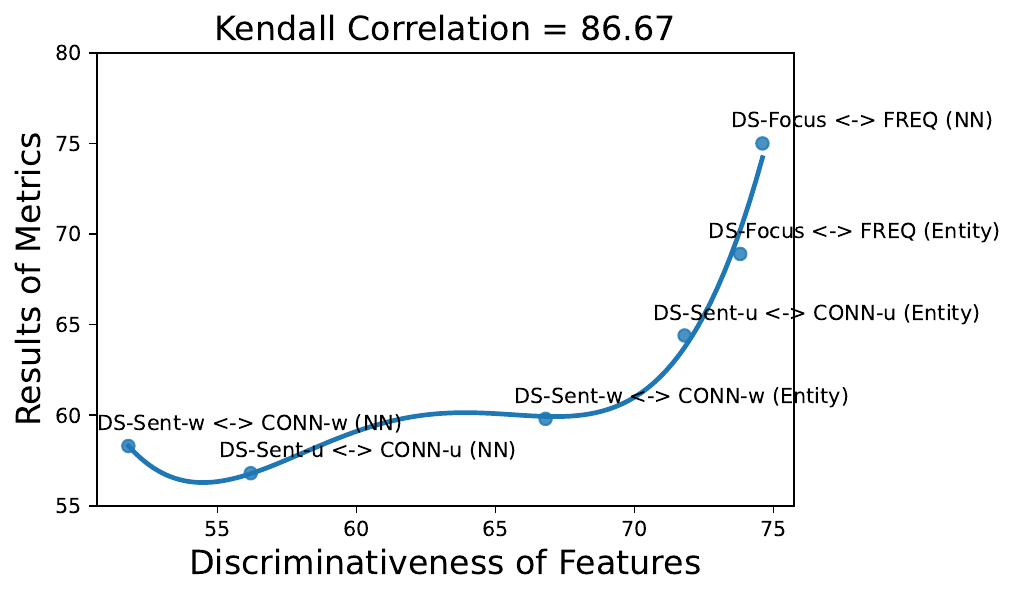}}  
 \caption{Correlations between the results of metrics and the discriminativeness of features on SUMMEval. Metric results are averaged across four rating aspects.}
 \label{fig:corr_regularity_results}  
\end{figure}

Figure \ref{fig:corr_regularity_results} shows that the discriminativeness of these features strongly correlate with the results of the DiscoScore variants, i.e., that the more discriminative the features are, the better the metrics perform. This attributes the superiority of a metric to the fact that the discourse feature can better separate hypothesis and reference.

From this, we can interpret the performance gaps between the DiscoScore variants, namely (i) \metric{DS-Focus} over \metric{DS-Sent}: given \emph{Focus Frequency} is more discriminative than \emph{Sentence Connectivity}, it is not surprising that \metric{DS-Focus} modeling discourse coherence with the former outperforms \metric{DS-Sent} modeling with the latter, and (ii)  
DS-Focus (NN) outperforms DS-Focus (Entity) because \emph{Frequency (NN)} can better separate hypothesis from reference than \emph{Frequency (Entity)}.

\paragraph{Analyses.} We provide analyses on the configuration of DiscoScore from three perspectives---see Appendix \ref{sec:analyses}: (i) the choice of BERT variants towards discourse- versus non-discourse BERT; (ii) the impact of adjacency matrices accounting for the interdependence between sentences and (iii) that we compare statistics- and alignment-based approaches to examine the best configuration for \metric{DS-Sent}. Our results show the advantages of adjacency matrices and statistics based approach, and that discourse BERT only helps for \metric{DS-Focus}.

\subsection{Document-level Machine Translation}

\paragraph{Correlation Results.} 

Table~\ref{tab:main-WMT20} (appendix) compares metrics on WMT20. We see that non-discourse metrics seem much better, but these results are not consistent to the discriminativeness of the discourse features---see Table~\ref{tab:statistics-WMT20} (appendix). For instance, in cs-en, the discourse features (Frequency and Connectivity) corresponding to \metric{DS-Focus} and \metric{DS-Sent} clearly separate hypothesis from reference due to the probability of $\mathcal{D}>0$ being over 70\%. However, both \metric{DS-Focus} and \metric{DS-Sent} correlate weakly with human rated adequacy. Recently, \citet{Markus:2021} provide justification to the inadequacy of the `adequacy' ratings, as `adequacy' sometimes cannot distinguish human from system translations and  correlates weakly with multiple aspects (e.g., fluency and accuracy). Thus, they re-annotate WMT20 with the MQM and pSQM rating schemes, which has been subsumed into the annotation guideline of the most recent WMT evaluation campaign~\cite{freitag-etal-2021-results}. Here, we perform an extra study on these ratings on both document- and system-levels. Note that system-level ratings are said to be the average of document-level ones in our setting. Table \ref{tab:mqm-ratings} (appendix) shows that \metric{DS-Sent} is much better than BARTScore on system level, surpassing it by 25 points in terms of MQM and 14 points in pSQM.

Overall, these results in MT are consistent with those in summarization, i.e., DiscoScore is strong on system levels for both tasks, but it cannot show gains on fine-grained levels. Section \ref{sec:mt-analyses} (appendix) show inter-correlations between metrics.

\section{Conclusions}
Given the recent growth in discourse based NLG systems, evaluation metrics targeting the assessment of text coherence are essential next steps for properly tracking the progress of these systems. 
Although there have been several attempts made towards discourse metrics, they all do not leverage strong contextualized encoders 
which have been held responsible for the recent success story of NLP. 
In this work, we introduced DiscoScore that uses BERT to model discourse coherence from two perspectives of readers' focus: (i) frequencies and semantics of foci and (ii) focus transitions over sentences used to predict interdependence between sentences. We find that BERT-based non-discourse metrics cannot address text coherence, even much worse than early feature-based discourse metrics invented a decade ago. We also find that the recent state-of-the-art BARTScore correlates weakly with human ratings on system level.
DiscoScore, on the other hand, performs consistently best in both single- and multi-reference settings, equally strong in coherence and several other aspects such as factual consistency, despite that we have not targeted 
them. More prominently, we provide understanding on the importance of discourse for evaluation metrics, and explain the superiority of one metric over another with simple features,
in line with recent work on explainability for evaluation metrics~\cite{kaster-etal-2021-global, fomicheva-etal-2021-eval4nlp}.

Scope for future research is huge, e.g., developing reference-free discourse metrics comparing source text to hypothesis, improving 
discourse metrics on fine-grained levels\footnote{
Recently, \citet{steen2022find} introduce a fine-grained evaluation setup to compute summary-level correlation, which performs computing over summaries not produced by multiple systems, but rather by a single system. This is because systems sometimes substantially differ in quality, which implies that involving multiple systems could result in inaccurate evaluation outcomes in the presence of system-level confounders.},
and ranking NLG systems 
via discourse metrics and rigorous
approaches~\cite{peyrard-etal-2021-better, DBLP:journals/corr/abs-2107-10821}.

\section{Impact and Limitation}
To our knowledge, we, for the first time, combine the elements of discourse and BERT representations to design an evaluation metric (DiscoScore) for text quality assessment in summarization and MT. While our experiments are conducted on English datasets, DiscoScore could adapt to many other languages in which references and foci are available. We believe that this work fosters future research on text generation systems endowed with the ability to produce well-formed texts in discourse.

However, we acknowledge several limitations of this work, which require further investigation in future. We now discuss them in the following:

\paragraph{Entity as Focus.} We follow the idea of \citet{mesgar-strube-2016-lexical} in the discourse community, which clusters nouns into entities based on their static word embeddings. Although simple, it sometimes helps for DiscoScore. However, alternatives aiming to produce better entities have not been explored in this work, e.g., replacing static with contextualized embeddings, and weighting entities by their occurrences in hypothesis/reference.

\paragraph{Weakness on Fine-Grained Assessment.} In summarization and MT, we show that our novel DiscoScore largely outperforms the current state-of-the-art BARTScore on system levels for both tasks, while it cannot show advantages on finer-grained levels such as document- and summary-levels.
This might be because modeling focus alone is insufficient to perform much more challenging, finer-grained 
assessment of text quality. Future work could also factor other discourse phenomena (e.g., discourse connectives and coreference) into the assessment of text coherence.

\section*{Acknowledgments}
We thank the anonymous reviewers for their thoughtful comments that greatly
improved the texts. 
This work has been supported by the German Research Foundation as part of the Research Training
Group Adaptive Preparation of Information from Heterogeneous Sources (AIPHES) at the Technische
Universit\"at Darmstadt under grant No. GRK 1994/1 and the Klaus Tschira Foundation, Heidelberg, Germany. Steffen Eger is funded by DFG Heisenberg grant EG 375/5-1.

\bibliography{anthology,wei}

\clearpage
\appendix
\section{Appendix}

\subsection{Evaluation Metrics}
\label{sec:details}
\paragraph{Non-discourse Metrics.} We consider the following non-discourse metrics.

\begin{itemize}
    \item BLEU~\cite{Papineni:2002} and ROUGE~\cite{Lin:2004} are precision- and recall-oriented metrics respectively, both of which measure n-gram overlap between a hypothesis and a reference. 
    
    \item BERTScore~\cite{DBLP:conf/iclr/ZhangKWWA20} and MoverScore~\cite{zhao-etal-2019-moverscore} are set-based metrics used to measure the semantic similarity between hypothesis and reference. BERTScore uses greedy alignment to compute the similarity between two sets of BERT-based word embeddings from hypothesis and from reference, while MoverScore uses optimal alignments based on Word Mover's Distance~\cite{kusner2015word} to do so.
    
    \item SBERT~\cite{DBLP:conf/emnlp/ReimersG19} fine-tunes BERT on the NLI datasets and uses pooling operations to produce sentence embeddings. We compute the cosine similarity between two sentence representations from hypothesis and from reference.  

    \item $S^{3}$-pyr and $S^{3}$-resp~\cite{peyrard-etal-2017-learning} are supervised metrics that 
    linearly combine ROUGE, JS-divergence and ROUGE-WE scores, trained on the TAC datasets with human annotated pyramid and responsiveness scores as supervision.
    
    \item BLEURT~\cite{sellam-etal-2020-bleurt} is another supervised metric that fine-tunes BERT on the concatenation of WMT datasets and synthetic data in the MT domain, with human judgment of translation quality as supervision.
    
    \item BARTScore~\cite{yuan2021bartscore} and PRISM~\cite{thompson-post-2020-automatic} depict sequence-to-sequence language models as metrics to compare hypothesis with reference. In reference-based settings, they both measure the likelihood that hypothesis and reference are paraphrases, but differ in the language models they rely on. PRISM has been based on a neural MT system trained from scratch on parallel data in MT, while BARTScore uses  BART~\cite{yuan2021bartscore} that has been fine-tuned on CNN/DailyMail~\cite{NIPS2015_afdec700}---which is parallel data in summarization. We use the `F-score' version of BARTScore as recommended in \citet{yuan2021bartscore}.

\end{itemize}

\paragraph{Discourse Metrics.} We consider the following discourse metrics (including ours and coherence models).
\begin{itemize}
    \item RC and LC~\cite{wong-kit-2012-extending} require neither source texts nor references and use lexical cohesion devices (e.g., repetition) within a hypothesis to predict text coherence. LC computes the proportion of words within hypothesis that are lexical cohesion devices, while RC computes the proportion of times that lexical cohesion devices appear in hypothesis.

    \item Entity Graph~\cite{guinaudeau-strube-2013-graph} and Lexical Graph~\cite{mesgar-strube-2016-lexical} are popular coherence models used to perform discourse tasks such as essay scoring, both of which introduce a graph with nodes as sentences and adjacency matrices as the connectivity between sentences. Here, we use the average of adjacency matrices from the hypothesis as the proxy of hypothesis coherence. While Entity Graph draws an edge between two sentences if both sentences have at least one noun in common, Lexical Graph draws an edge if two sentences have a pair of similar words in common, i.e., the cosine similarity between their embeddings greater than a threshold.
    
    \item Lexical Chain~\cite{gong2015document} 
    extracts multiple lexical chains from hypothesis and from reference. Each word is associated to a lexical chain if a word appears in more than one sentence. A lexical chain contains a set of sentence positions in which a word appears. Finally, the metric performs soft matching to measure lexical chain overlap between hypothesis and reference.
    
    \item FocusDiff and SentGraph are the two variants of DiscoScore, which use BERT to model semantics and coherence of readers' focus in hypothesis and reference. In particular, FocusDiff measures the difference between a common set of foci in hypothesis and reference in terms of semantics and frequency, while SentGraph measures the semantic similarity between two sets of sentence embeddings from hypothesis and reference---which are aggregated according to the number of foci shared across sentences and the distance between sentences.
\end{itemize}

\subsection{Datasets}
\label{sec:datasets}
We outline two datasets in summarization, and one in document-level MT. 
\paragraph{Text Summarization.}
While DUC\footnote{\url{https://duc.nist.gov/data.html}} and TAC\footnote{\url{https://tac.nist.gov/data/}} datasets with human rated summaries, constructed one decade ago, were the standard benchmarks for comparing evaluation metrics in summmarization,
they collect summaries only from extractive summarization systems.
In the last few years, abstractive systems have become popular; however, little is known how well metrics judge them. Recently, several datasets based on CNN/DailyMail have been constructed to address this. 
For instance, SummEval~\cite{fabbri2021summeval}, REALSumm~\cite{Bhandari-2020-reevaluating}, XSum~\cite{maynez_acl20} and FEQA~\cite{durmus-etal-2020-feqa} all collect summaries from both extractive and abstractive systems, but differ in the aspects human experts rate summaries. In this work, we consider the following two complementary summarization datasets.

\begin{itemize}
    \item SummEval has been constructed in multiple-references settings, i.e., that each hypothesis is associated to multiple references. It contains human judgments of summary coherence, factual consistency, fluency and relevance. We only consider abstractive summaries as they have little lexical overlap with references.
    
    \item NeR18~\cite{grusky-etal-2018-newsroom}, in contrast, has been constructed in single-reference settings. It contains human judgments of summary coherence, fluency, informativeness and relevance. As majority of summaries are extractive, we include both extractive and abstractive for the inclusive picture. 
\end{itemize}

\paragraph{Document-level Machine Translation.}
As document-level human ratings in MT are particularly laborious, hardly ever have there been MT datasets directly addressing them.  First attempts suggested to use the average of much cheaper sentence-level ratings as a document score for comparing document-level metrics~\cite{comelles-etal-2010-document, wong-kit-2012-extending, gong2015document}. However, human experts were asked to rate sentences in isolation within a document. Thus, human ratings at both sentence and document levels cannot reflect inter-sentence coherence. Recently, the WMT20 workshop~\cite{mathur-etal-2020-results} asks humans to rate each sentence translation in the document context, and follows the previous idea of `average' to yield document scores.

In this work, we use the WMT20 dataset with `artificial' document-level ratings. Note that WMT20 comes with two issues: (i) though sentences are rated in the document context, averaging sentence-level ratings may zero out negative effects of incoherent elements on document level and (ii) unlike SummEval and NeR18, WMT20 only contains human judgment of translation \emph{adequacy} (which may subsume multiple aspects), not \emph{coherence}. 
 
For simplicity, we exclude system and reference translations with lengths greater than 512---the number of tokens at maximum allowed by BERT, as only a small portion of instances is over the token limit. Note that it is effortless to replace BERT with Longformer~\cite{Beltagy2020Longformer} to deal with longer documents for DiscoScore.

\subsection{Analyses on Text Summarization}
\label{sec:analyses}
\insertTableLM
\insertTableAblation
\insertTableAggregation
\insertTableSummaryLevel

\paragraph{Choice of BERT Variants.} Table~\ref{tab:LM} compares the impact of three BERT variants on DiscoScore. 
Conpono, referred to as a discourse BERT, has finetuned BERT with a novel discourse-level objective regarding sentence ordering. While strong on discourse evaluation benchmarks~\cite{chen-etal-2019-evaluation},
Conpono is not always helpful, e.g., BERT-NLI is better for \metric{DS-Sent}. These results suggest the best configuration for DiscoScore.

\paragraph{Impact of Sentence Connectivity.} Table~\ref{tab:ablation} shows an ablation study on the use of sentence connectivity. Aggregating sentence embeddings with our adjacency matrices (see Eq.\ref{eq:sent-graph}) helps considerably. This confirms the usefulness of aggregation from which we include coherence signals in sentence embeddings.

\paragraph{SentGraph Variants.}Table~\ref{tab:greedy-optimal} compares three \metric{DS-Sent} variants as to how we measure the distance between two sets of sentence embeddings from hypothesis and reference. In particular, we refer to  
BERTScore~\cite{DBLP:conf/iclr/ZhangKWWA20} as a `greedy align' mechanism used to compute the similarity between two sets of sentence embeddings. As for `optimal align', we use MoverScore~\cite{zhao-etal-2019-moverscore} to do so. 
While the two alignments directly measure the distance between the two sets, the simple statistics, i.e., mean-max-min-sum, derives a graph embedding from each set and computes the cosine similarity between two graph embeddings. We see that the `statistics' wins by a big margin, and thus adopt this \metric{DS-Sent} variant in all setups.

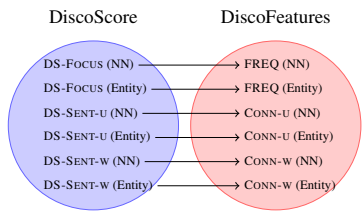
\begin{figure}
\centering
\tiny
\scalebox{0.8}{
\begin{tikzpicture}[al/.style={align=right}]
    \filldraw[fill=blue!20, draw=blue!60] (-1.5,-0.3) circle (1.4cm);
    \filldraw[fill=red!20, draw=red!60] (1.5, -0.3) circle (1.4cm);
    
    \node at (-1.5,1.5) {\footnotesize DiscoScore};
    \node at (1.5,1.5) {\footnotesize DiscoFeatures};

    \node[anchor=west] (x1) at (-2.4,0.7) {\metric{DS-Focus} (NN)};
    \node[anchor=west] (x2) at (-2.4,0.3) {\metric{DS-Focus} (Entity)};
    \node[anchor=west] (x3) at (-2.4,-0.1) {\metric{DS-Sent-u} (NN)};
    \node[anchor=west] (x4) at (-2.4,-0.5) {\metric{DS-Sent-u} (Entity)};
    \node[anchor=west] (x5) at (-2.4,-0.9) {\metric{DS-Sent-w} (NN)};
    \node[anchor=west] (x6) at (-2.4,-1.3) {\metric{DS-Sent-w} (Entity)};
    \node[anchor=west] (y1) at (0.9,0.7) {\metric{FREQ} (NN)};
    \node[anchor=west] (y2) at (0.9,0.3) {\metric{FREQ} (Entity)};
    \node[anchor=west] (y3) at (0.9,-0.1) {\metric{Conn-u} (NN)};
    \node[anchor=west] (y4) at (0.9,-0.5) {\metric{Conn-u} (Entity)};
    \node[anchor=west] (y5) at (0.9,-0.9) {\metric{Conn-w} (NN)};
    \node[anchor=west] (y6) at (0.9,-1.3) {\metric{Conn-w} (Entity)};
    \draw[->] (x1) -- (y1);
    \draw[->] (x2) -- (y2);
    \draw[->] (x3) -- (y3);
    \draw[->] (x4) -- (y4);
    \draw[->] (x5) -- (y5);
    \draw[->] (x6) -- (y6);

\end{tikzpicture}
}
\caption{Links between the DiscoScore variants and discourse features. 
}
\label{fig:mapping}
\end{figure}

\subsection{Analyses on MT}
\label{sec:mt-analyses}
\paragraph{Correlation between Metrics.}

\begin{figure*}
\begin{minipage}{0.24\textwidth}  
	\centerline{\includegraphics[width=\linewidth]{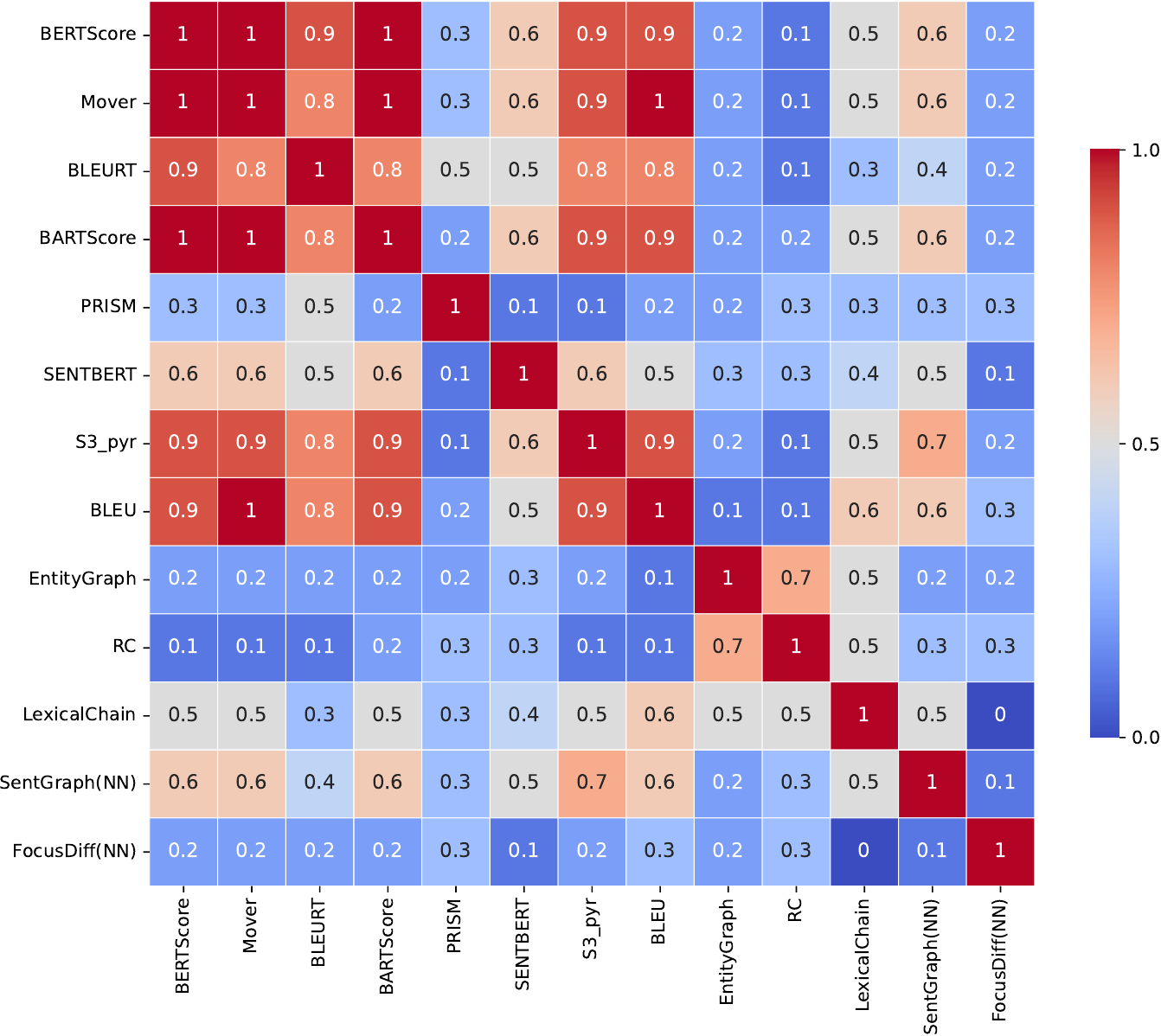}} 
\end{minipage} 
\begin{minipage}{0.24\textwidth}  
	\centerline{\includegraphics[width=\linewidth]{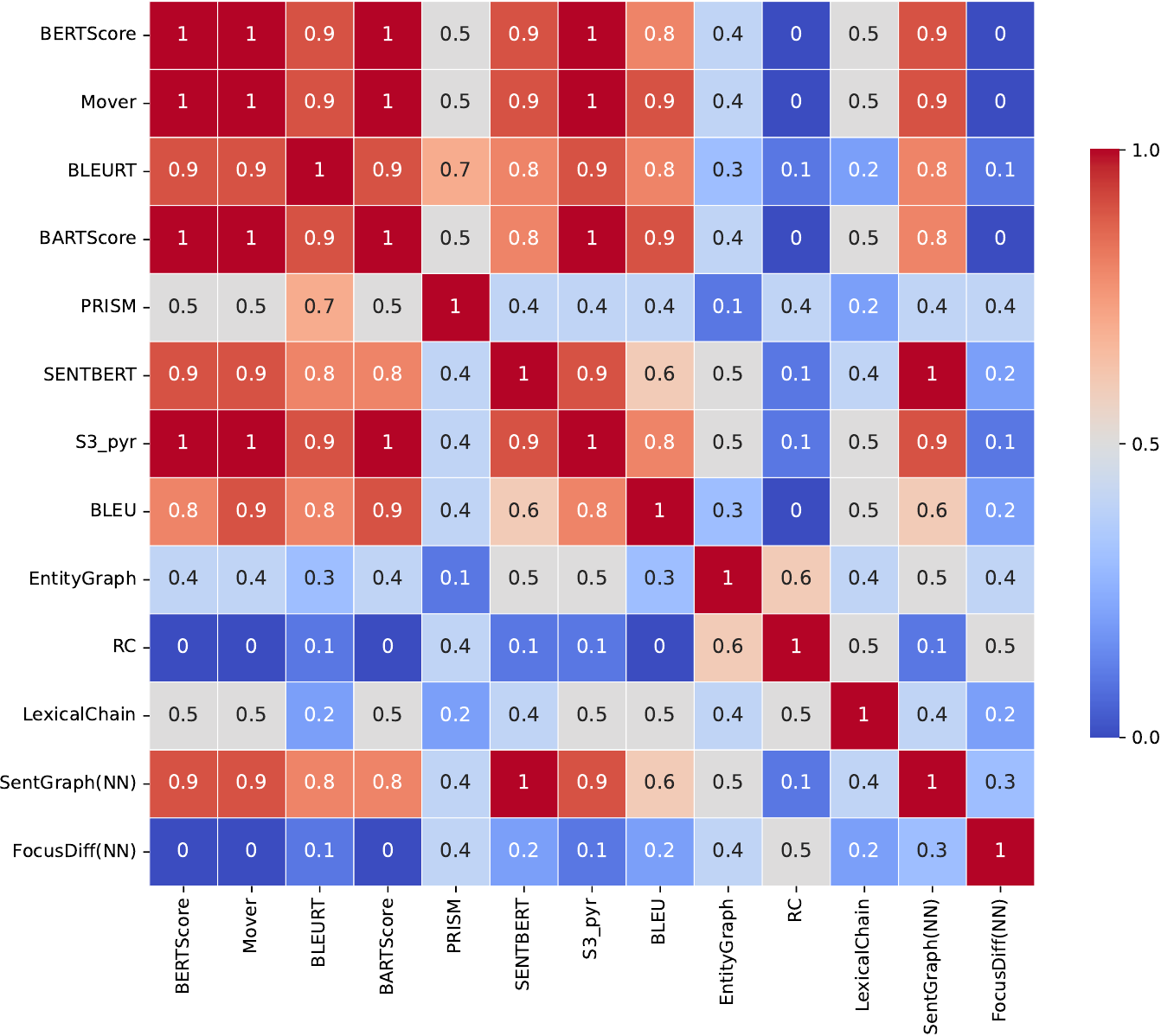}}  
\end{minipage} 
\begin{minipage}{0.24\textwidth}  
	\centerline{\includegraphics[width=\linewidth]{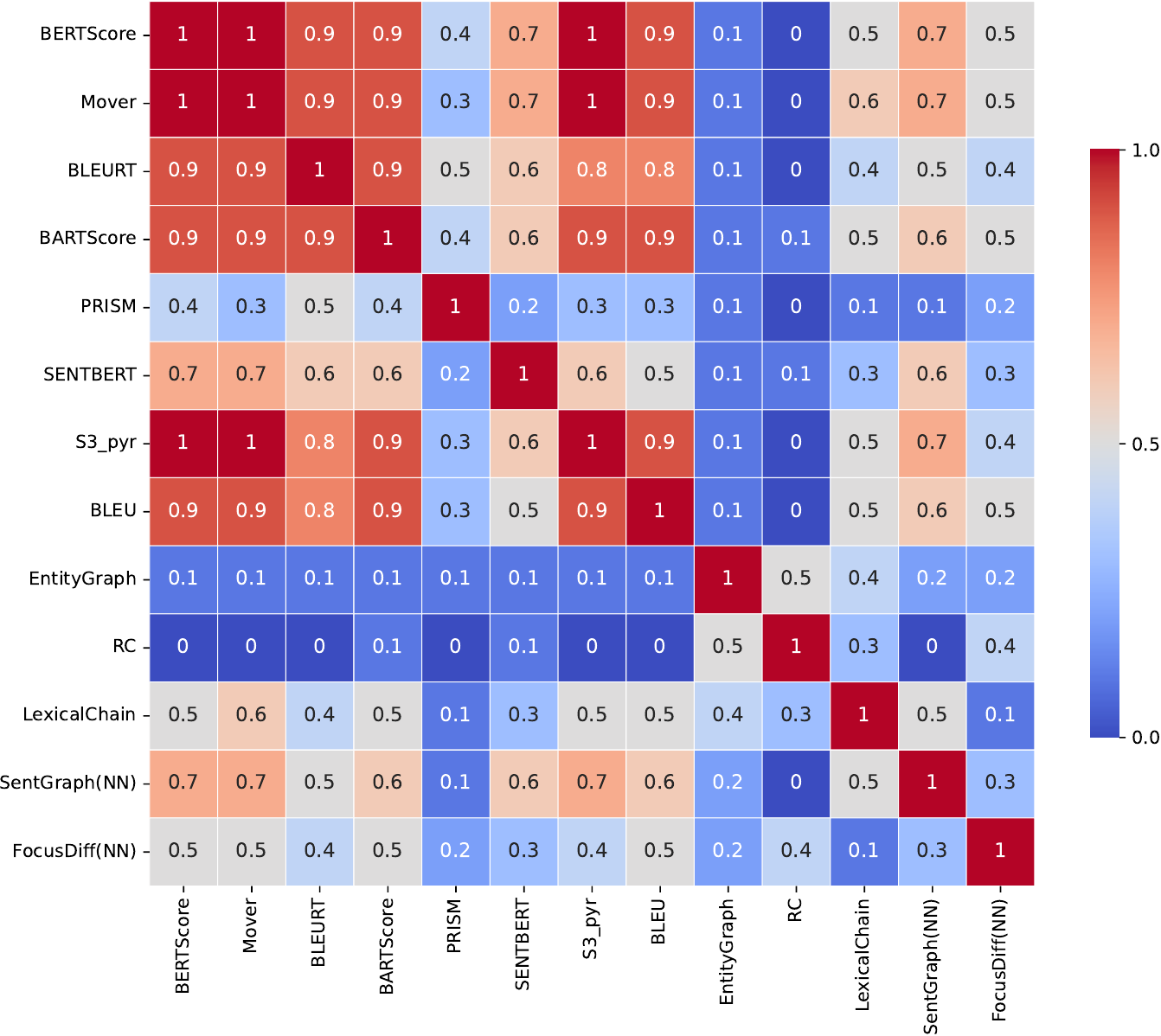}}  
\end{minipage} 
\begin{minipage}{0.24\textwidth}  
	\centerline{\includegraphics[width=\linewidth]{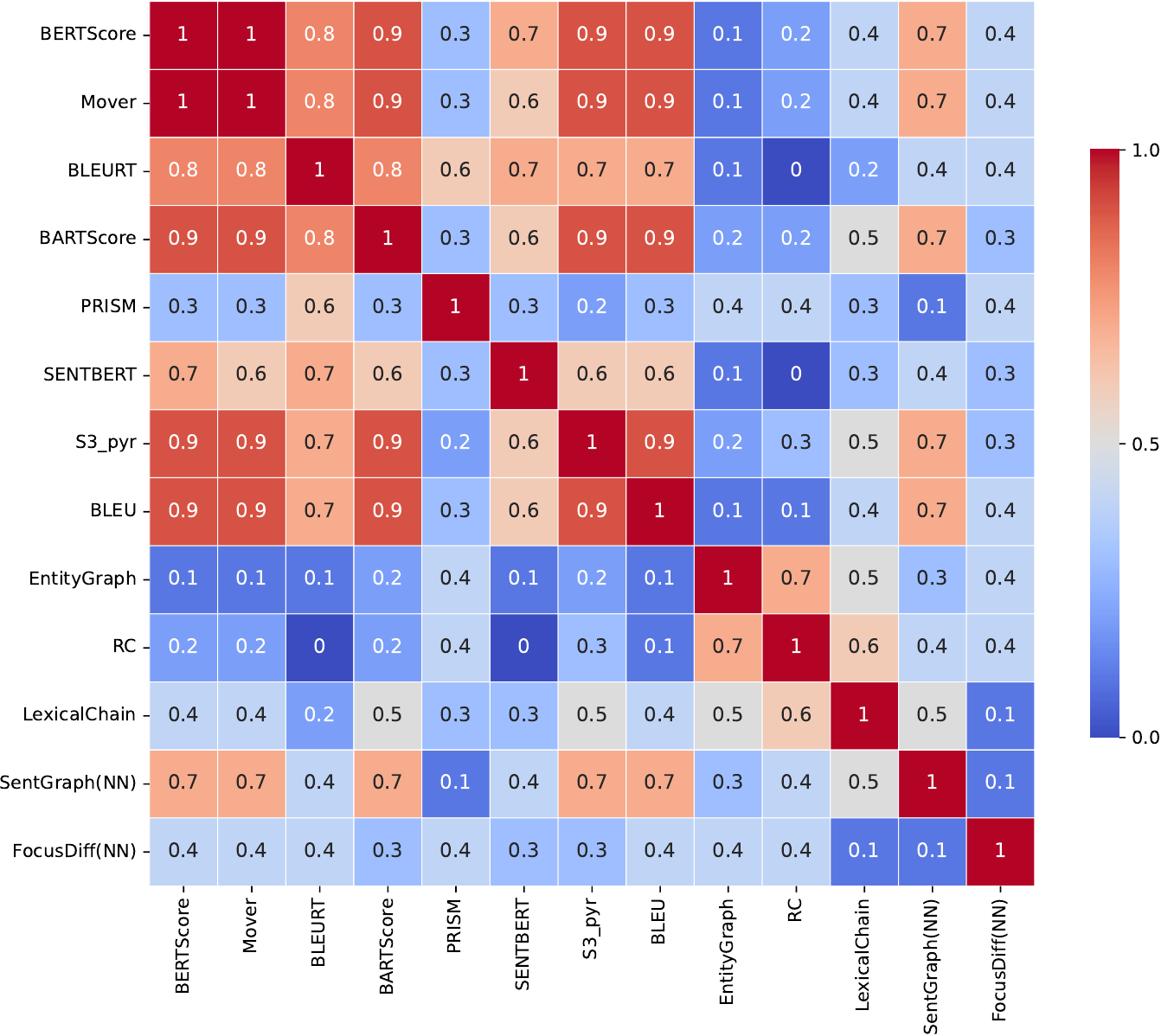}}
\end{minipage} 
\caption{Pearson Correlations between  metrics on WMT20 in cs-en, de-en, ja-en and ru-en (from left to right).
}
\label{fig:corr-remaining-lang}
\end{figure*}

Figure~\ref{fig:corr-remaining-lang} shows inter-correlations between metrics on WMT20 across languages. Overall, correlations are mostly high between non-discourse metrics, much weaker between discourse and non-discourse metrics---which confirms the orthogonality of them in that they rate translations in different aspects. We note that \metric{DS-Focus} has the lowest correlations with all other metrics. For instance, \metric{DS-Focus} is almost orthogonal to BERTScore and MoverScore. We investigated whether combining them receives additive gains. We find that a combination of \metric{DS-Focus} and BERTScore (or MoverScore) provides little help in correlation with adequacy.

\begin{table}
    \footnotesize
    \centering
    \begin{tabular}{l | rrrr}
    \toprule
    & \multicolumn{2}{c}{\textbf{Sys-level}} & \multicolumn{2}{c}{\textbf{Doc-level}}\\
     \textbf{Metrics} & \textbf{MQM} & \textbf{pSQM} & \textbf{MQM} & \textbf{pSQM}\\
    \midrule
    BARTScore & 45.57 & 55.50 & \textbf{34.90} & \textbf{28.96} \\
    \metric{*DS-Focus} (NN) & 42.12 & 40.89 & 19.10 & 9.98\\
    \metric{DS-Sent-u} (NN) & \textbf{70.77} & \textbf{69.74} & 19.98 & 14.49\\
    \bottomrule
    \end{tabular}
    \caption{Document-level Kendall and system-level Pearson correlations between metrics and MQM/pSQM ratings on WMT20 in Chinese-to-English---which is the only language pair with such ratings in reference-based settings. \metric{*DS-Focus} (NN) excludes focus that occurs only once in hypothesis/reference.
    }

    \label{tab:mqm-ratings}
\end{table}

\insertTableNeR

\begin{table*}[!htbp]
    \footnotesize
    \setlength{\tabcolsep}{3pt}
      \centering
        \begin{tabular}{l|l| rrrrr}
            \toprule
            \textbf{Settings} & \textbf{Metrics} & \textbf{Coherence} & \textbf{Consistency} & \textbf{Fluency} & \textbf{Relevance} & \textbf{Average} \\
            \midrule
            \multirow{6}{*}{$m(\mathrm{hyp}, \mathrm{ref})$}
            &BARTScore (max) & \textbf{78.79} & 48.48 & 63.64 & 72.73 & 65.91\\
            & BARTScore (original) & 60.61 & 36.36 & 45.45 & 48.48 & 47.73\\
            \cmidrule{2-7}
            &FocusDiff (NN) & \textbf{75.76} & \textbf{63.64} & \textbf{78.79} & \textbf{81.82} & \textbf{75.00} \\
            &FocusDiff (Entity) & 69.70 & 57.58 & 72.73 & 75.76 & 68.94 \\ 
            &SentGraph-u (NN) & 48.48 & 54.55 & 63.64 & 60.61 & 56.82 \\
            &SentGraph-u (Entity) & 54.55 & 60.61 & 75.76 & 66.67 & 64.39 \\ 
        	\bottomrule  
        \end{tabular}
   
    \caption{System-level Kendall correlations between metrics and human ratings on SUMMEval in multi-reference settings. BARTScore (original) compares a hypothesis with one reference at a time, and takes the average of evaluation scores as a final score, while BARTScore (max) takes the maximum score.
    \label{tab:main-extra-SUMMEval} 
    }
\end{table*}

\begin{table*}[t!]
    \footnotesize
    \centering
    \begin{tabular}{l | rr|rrrrr}
    \toprule
    & & & \multicolumn{4}{c}{\textbf{WMT20}}\\
    & \textbf{SUMMEval} & \textbf{NeR18} & 
    \textbf{cs-en} &\textbf{de-en} & \textbf{ja-en} & \textbf{ru-en}\\
    \midrule
    Number of references & 11 & 1& 1 & 1 & 1 & 1\\
    Number of systems & 12 & 7 & 13 & 14 & 11 & 13\\
    Number of hypothesis per system & 100 & 60 & 102 & 118 & 80 & 91\\
    Number of sentences per hypothesis & 3.13 & 1.90 & 15.21 &13.84 & 11.29 & 9.46\\
    Average number of foci in hypothesis & 15.18 & 12.85 & 62.01 & 56.68 & 57.09 & 44.99\\
    Average number of `good foci' in hypothesis & 2.47 & 2.56& 13.16 & 13.37 & 15.07 & 9.95\\
    Percent of hypotheses with `good foci' & 80.50\% & 43.80\%& 100\% & 98.60\% & 100\% & 100\%\\
    \bottomrule
    \end{tabular}
    \caption{Characteristics of summarization and MT datasets. `good foci' denotes a focus appearing more than once in hypothesis. The more often a focus appears, the stronger the discourse signals are.}

    \label{tab:statistics}
\end{table*}

\insertTableEnsemble

\begin{figure*}[htb]
    \centering 
\begin{subfigure}{0.3\textwidth}
  \includegraphics[width=\linewidth]{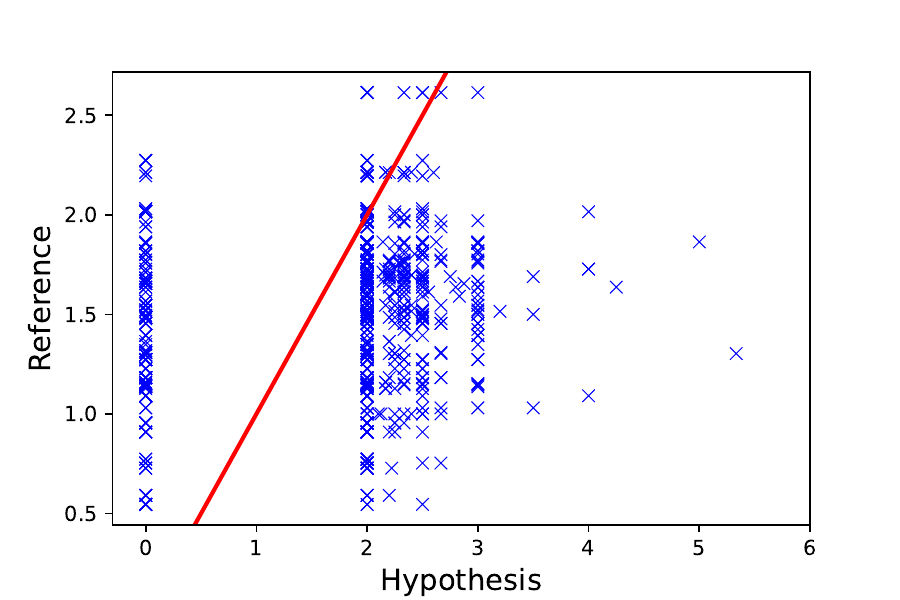}
  \caption{Focus Frequency (NN)}
\end{subfigure}\hfil 
\begin{subfigure}{0.3\textwidth}
  \includegraphics[width=\linewidth]{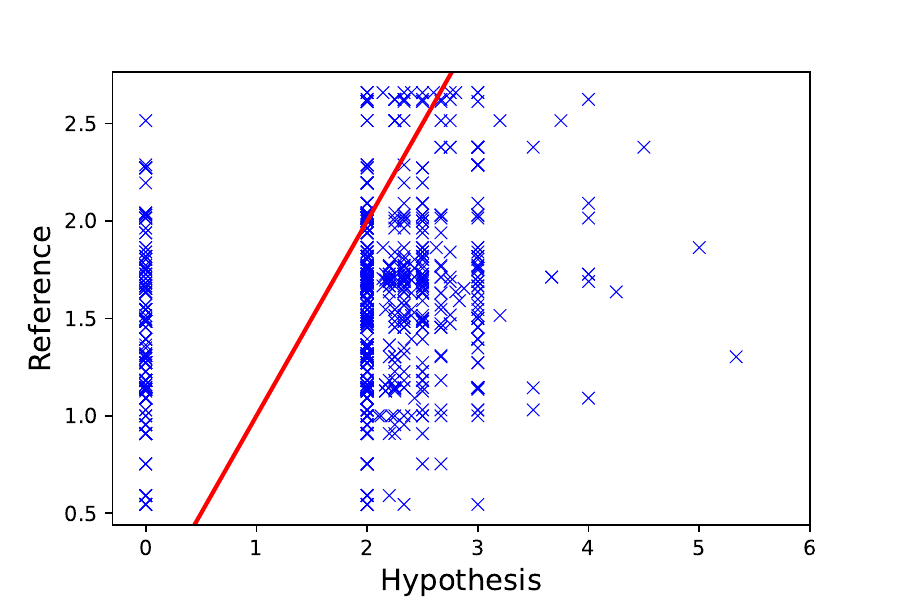}
  \caption{Focus Frequency (Entity)}
\end{subfigure}\hfil 
\begin{subfigure}{0.3\textwidth}
  \includegraphics[width=\linewidth]{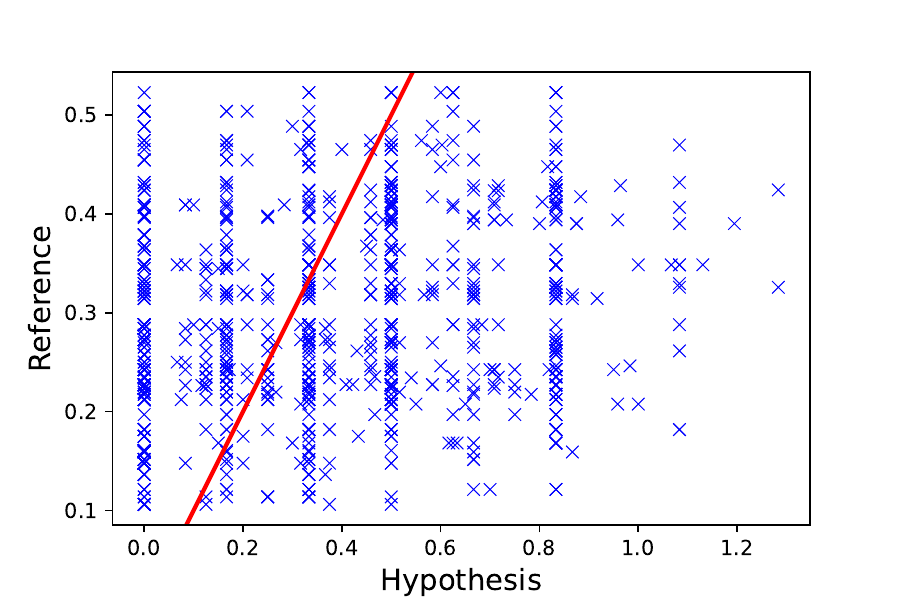}
  \caption{Connectivity-u (NN)}
\end{subfigure}

\medskip
\begin{subfigure}{0.3\textwidth}
  \includegraphics[width=\linewidth]{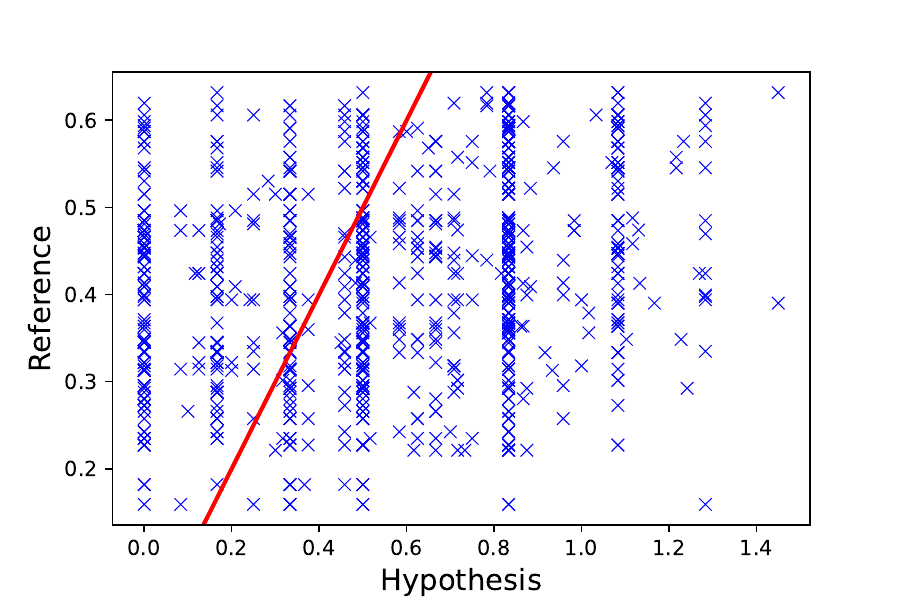}
  \caption{Connectivity-u (Entity)}
\end{subfigure}\hfil 
\begin{subfigure}{0.3\textwidth}
  \includegraphics[width=\linewidth]{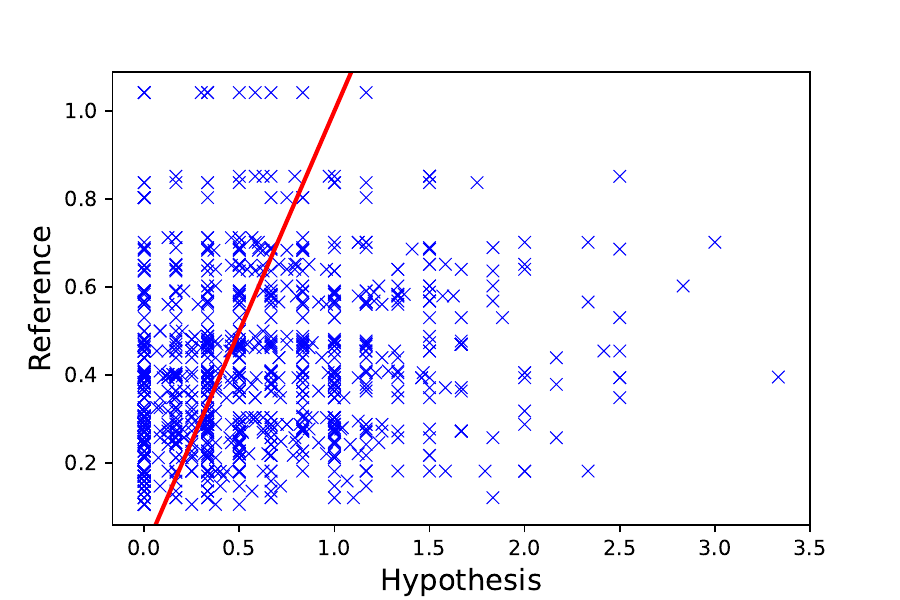}
  \caption{Connectivity-w (NN)}
\end{subfigure}\hfil 
\begin{subfigure}{0.3\textwidth}
  \includegraphics[width=\linewidth]{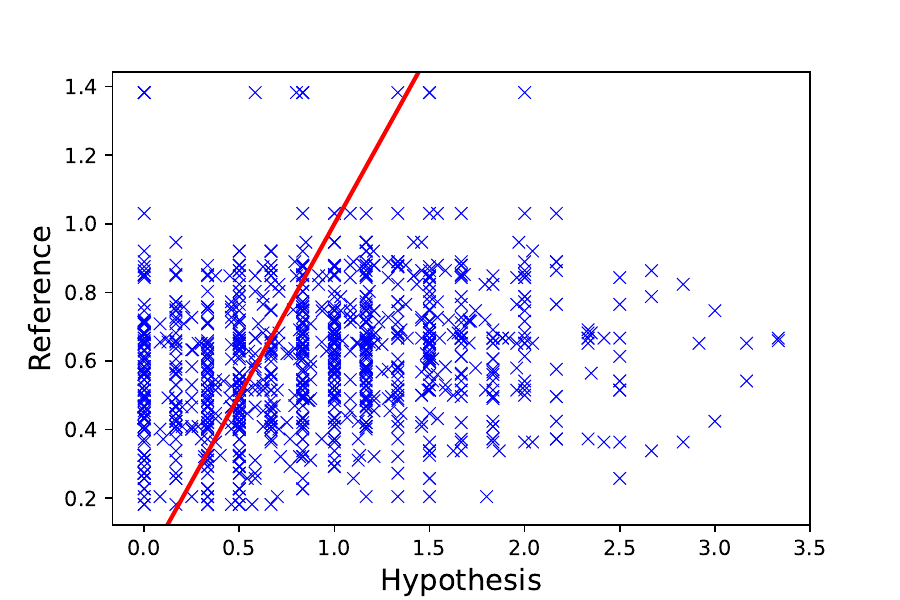}
  \caption{Connectivity-w (Entity)}
\end{subfigure}
\caption{Distribution of discourse features over hypothesis and reference on SUMMEval.}
\label{fig:dist-all}
\end{figure*}

\begin{table*}[!htb]
    \centering
    \setlength{\tabcolsep}{2.5pt}
    \footnotesize
    \begin{tabular}{l|ccc|ccc|ccc|ccc}
    \toprule
        & \multicolumn{3}{c}{\textbf{cs-en}} & \multicolumn{3}{c}{\textbf{de-en}} & \multicolumn{3}{c}{\textbf{ja-en}} & \multicolumn{3}{c}{\textbf{ru-en}}\\
        \textbf{DiscoFeatures} & $\mathcal{D}>0$ & $\mathcal{D}=0$ & $\mathcal{D}<0$ 
        & $\mathcal{D}>0$ & $\mathcal{D}=0$ & $\mathcal{D}<0$ 
        & $\mathcal{D}>0$ & $\mathcal{D}=0$ & $\mathcal{D}<0$
        & $\mathcal{D}>0$ & $\mathcal{D}=0$ & $\mathcal{D}<0$\\ 
        \midrule
        Frequency (NN) & 74.18 & 2.00 & 23.82 & 57.38 & 9.65 & 32.97 & 53.04 & 2.63 & 44.33 & 52.77 & 7.31 & 39.92\\
        
        Frequency (Entity) & 76.17 & 1.76 & 22.07 & 59.74 & 8.38 & 31.88 & 52.38 & 1.48 & 46.14 & 53.61 & 7.31 & 39.08\\
        
        Connectivity-u (NN) & 78.05 & 0.35 & 21.60 & 63.11 & 8.29 & 28.60 & 59.61 & 5.25 & 35.14 & 52.04 & 10.03 & 37.93 \\
        
        Connectivity-u (Entity) & 79.46 & 0.35 & 20.19 & 62.02 & 8.20 & 29.78 & 59.44 & 5.09 & 35.47 & 52.87 & 9.40 & 37.72 \\
        Connectivity-w (NN) & 77.93 & 0.24 & 21.83 & 64.85 & 4.64 & 30.51 & 59.12 & 0.49 & 40.39 & 59.98 & 5.12 & 34.90 \\
        Connectivity-w (Entity) & 80.40 & 0.23 & 19.37 & 63.48 & 4.73 & 31.79 & 60.76 & 0.33 & 38.91 & 60.82 & 4.60 & 34.58\\

        \bottomrule
    \end{tabular}
    \caption{Statistics of discourse features on WMT20. $\mathcal{D}>0$ denotes the percent of `reference-hypothesis' pairs for which $\mathcal{R}(\mathrm{ref})>\mathcal{R}(\mathrm{hyp})$ with $\mathcal{R}$ as any one of these features,
    similarly for the definitions of $\mathcal{D}=0$ and $\mathcal{D}<0$. 
    We exclude the pairs for which hypothesis and reference are the exact same.
    }
    \label{tab:statistics-WMT20}
\end{table*}

\begin{table*}[!htbp]
    \setlength{\tabcolsep}{6pt}
      \centering
      \footnotesize
        \begin{tabular}{l|l| rrrrr}
            \toprule
            & & \multicolumn{5}{c}{\textbf{Direct Assessment (Adequacy)}} \\
            \textbf{Settings} & \textbf{Metrics} & \textbf{cs-en} & \textbf{de-en} & \textbf{ja-en} & \textbf{ru-en} & \textbf{Average}\\ 
            \midrule
            \multirow{11}{*}{$m(\mathrm{hyp}, \mathrm{ref})$}
            &\multicolumn{4}{l}{\textbf{Non-discourse metrics}}\\
            \cmidrule{2-7}
            &BLEU & 7.44 & 57.52 & 41.48 & 10.74 & 29.30\\
            &BERTScore & 10.82 & 60.38 & \textbf{46.95} & 13.08 & 32.81 \\
            &MoverScore & \textbf{15.40} & \textbf{61.69} & 42.12 & 13.78 & \textbf{33.25}\\
            &BARTScore & 10.82 & 60.26 & 46.30 & 14.95 & 33.09\\
            &PRISM & 8.64 & 58.83 & 32.48 & 15.42 & 28.84\\
            &SBERT & 13.20 & 55.26 & 33.44 & 10.04 & 27.99\\
            &BLEURT & 12.01 & 58.83 & 37.94 & \textbf{18.22} & 31.75\\
            &$S^3$-pyr & 6.25 & 58.83 & 42.44 & 13.78 & 30.33\\
            &$S^3$-resp & 5.85 & 58.59 & \textbf{47.26} & 14.71 & 31.61\\
            \midrule
            \multirow{6}{*}{$m(\mathrm{hyp})$}
            & \multicolumn{4}{l}{\textbf{Discourse metrics}}\\
            \cmidrule{2-7}
            &RC & 5.85 & 7.19 & \textbf{8.68} & 9.34 & 7.77\\
            &LC & \textbf{9.23} & 1.72 & 3.53 & 6.07 & 5.14\\
            &Entity Graph & 5.06 & \textbf{43.24} & 3.53 & 10.51 & 15.59\\
            &Lexical Graph & 2.28 & \textbf{43.60} & 5.14 & \textbf{13.55} & \textbf{16.15}\\
            \midrule
            \multirow{9}{*}{$m(\mathrm{hyp}, \mathrm{ref})$}
            & \multicolumn{4}{l}{\textbf{Discourse metrics}}\\
            \cmidrule{2-7}
            &Lexical Chain &  21.54 & 35.15 & 15.11 & 16.12 & 21.99\\
            &FocusDiff (NN) & 7.64 & 33.13 & 19.29 & 2.57 & 15.66\\
            &FocusDiff (Entity) & 6.45 & 33.73 & 19.94 & 1.64 & 15.44\\

            &SentGraph-u (NN) & 7.64 & 57.16 & 39.22 & 18.22 & \textbf{30.56}\\
            &SentGraph-u (Entity) & 7.65 & 57.17 & 39.23 & 18.22 & \textbf{30.57}\\

            &SentGraph-w (NN) & 7.65 & 57.18 & 39.22 & 18.21 & \textbf{30.57}\\
            &SentGraph-w (Entity) &7.65 & 57.17 & 39.23 & 18.22 & \textbf{30.57}\\
        	\bottomrule  
        \end{tabular}
      
    \caption{Document-level Kendall correlations between metrics and human rated translation quality on WMT20. 
    \label{tab:main-WMT20} 
    }
\end{table*}

\end{document}